\documentclass{bmvc2k}

\usepackage{bm}
\usepackage{amsmath}
\usepackage{amssymb}
\usepackage{colortbl}
\usepackage{booktabs}
\usepackage{multirow}
\usepackage{makecell}
\usepackage{hyperref}
\usepackage{cleveref}
\usepackage{graphicx}
\crefname{section}{Sec.}{Secs.}
\crefname{table}{Tab.}{Tabs.}
\crefname{figure}{Fig.}{Figs.}
\definecolor{tabfirst}{rgb}{1, 0.7, 0.7} 
\definecolor{tabsecond}{rgb}{1, 0.85, 0.7} 
\definecolor{tabthird}{rgb}{1, 1, 0.7} 

\title{Drone-assisted Road Gaussian Splatting with Cross-view Uncertainty}

\addauthor{Saining Zhang}{SAINING001@e.ntu.edu.sg}{1,2\footnotemark[1]}
\addauthor{Baijun Ye}{yebaijun52@gmail.com}{1,3\footnotemark[1]}
\addauthor{Xiaoxue Chen}{chenxx21@mails.tsinghua.edu.cn}{1}
\addauthor{Yuantao Chen}{yuantaochen973@gmail.com}{1}
\addauthor{Zongzheng Zhang}{zzongzheng0918@gmail.com}{1}
\addauthor{Cheng Peng}{1120211642@bit.edu.cn}{1,4}
\addauthor{Yongliang Shi}{shiyongliang@air.tsinghua.edu.cn}{1}
\addauthor{Hao Zhao}{zhaohao@air.tsinghua.edu.cn}{1\footnotemark[2]}

\addinstitution{
 Institute for AI Industry Research (AIR),\\
 Tsinghua University,\\
 Beijing, China
}
\addinstitution{
 College of Computing and Data Science,\\
 Nanyang Technological University,\\
 Singapore
}
\addinstitution{
 IIIS,\\
 Tsinghua University,\\
 Beijing, China
}
\addinstitution{
 School of Computer Science
and Technology,\\
 Beijing Institute of Technology,\\
 Beijing, China
}

\runninghead{ZHANG ET AL.}{UC-GS}

\begin{document}
\footnotetext[1]{Equal contribution}
\footnotetext[2]{Corresponding author}
\maketitle

\begin{abstract}

Robust and realistic rendering for large-scale road scenes is essential in autonomous driving simulation. Recently, 3D Gaussian Splatting (3D-GS) has made groundbreaking progress in neural rendering, but the general fidelity of large-scale road scene renderings is often limited by the input imagery, which usually has a narrow field of view and focuses mainly on the street-level local area. Intuitively, the data from the drone's perspective can provide a complementary viewpoint for the data from the ground vehicle's perspective, enhancing the completeness of scene reconstruction and rendering. However, training naively with aerial and ground images, which exhibit large view disparity, poses a significant convergence challenge for 3D-GS, and does not demonstrate remarkable improvements in performance on road views. In order to enhance the novel view synthesis of road views and to effectively use the aerial information, we design an uncertainty-aware training method that allows aerial images to assist in the synthesis of areas where ground images have poor learning outcomes instead of weighting all pixels equally in 3D-GS training like prior work did. We are the first to introduce the cross-view uncertainty to 3D-GS by matching the car-view ensemble-based rendering uncertainty to aerial images, weighting the contribution of each pixel to the training process. Additionally, to systematically quantify evaluation metrics, we assemble a high-quality synthesized dataset comprising both aerial and ground images for road scenes. Through comprehensive results, we show that: (1) Jointly training aerial and ground images helps improve representation ability of 3D-GS when test views are shifted and rotated, but performs poorly on held-out road view test. (2) Our method reduces the weakness of the joint training, and out-performs other baselines quantitatively on both held-out tests and scenes involving view shifting and rotation on our datasets. (3) Qualitatively, our method shows great improvements in the rendering of road scene details, as shown in \cref{fig:teaser}. The code and data for this work will be released at \url{https://github.com/SainingZhang/UC-GS}.

\end{abstract}

\section{Introduction}
\label{sec:intro}



Autonomous driving simulation serves as a critical platform for scaling up to real-world deployment. Realistic rendering and novel view synthesis (NVS) for large-scale road scenes have become increasingly important in autonomous driving simulation, as they enable the synthesis of high-quality training and testing data at a significantly lower cost compared to using real-world data. This capability also benefits a wide range of applications, including digital cities \cite{li2024nerf,ye2024blending}, virtual reality \cite{xu2023vr}, and embodied AI \cite{qiu2024learning}. 



Recently, NeRF \cite{mildenhall2021nerf,ost2021neural,wu2023mars,turki2023suds,yuan2024presight, wei2024editable} has greatly enhanced fidelity of NVS by parameterizing the 3D scene as implicit neural fields but suffers from slow rendering due to exhaustive per-pixel ray sampling process especially in large-scale road scenes. 3D Gaussian Splatting (3D-GS) \cite{kerbl20233d} achieves real-time rendering by rasterizing the learnable Gaussian primitives.

\begin{figure*}
\begin{center}
\includegraphics[width=1.0\linewidth]{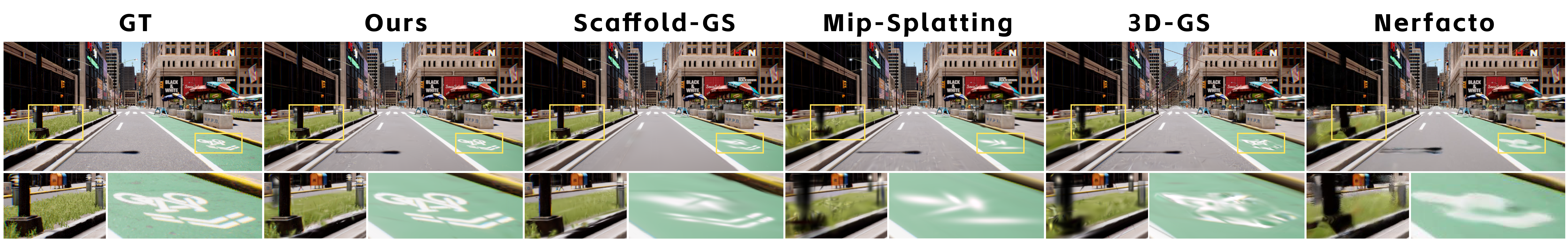}
\end{center}
\vspace{-5mm}
   \caption{Qualitative results of our Drone-assisted Road Gaussian Splatting with Cross-view Uncertainty and several baseline methods. The dataset is 1.6m test set of New York City. The quality improvement is highlighted by boxes.}
\endcenter
\label{fig:teaser}
\vspace{-5mm}
\end{figure*}

However, the rendering quality for both NeRF and 3D-GS is highly dependent on the input views.
On the contrary, current road scene datasets, such as KITTI \cite{geiger2013vision} or nuScenes \cite{caesar2020nuscenes}, only contain car-view images, which are limited by the field of view and focus on the street-level local areas.
The most related work may be MatrixCity \cite{li2023matrixcity}, which offers both aerial and street-level city views, but the aerial imagery's high altitude limits its ability to capture fine-grained road details. This mismatch in scale and granularity between global aerial views and local street views makes MatrixCity unsuitable for drone-assisted road scene synthesis.

To address the aforementioned issues, we dedicate to establish a new paradigm for \textbf{drone-assisted road scene synthesis}, aiming to overcome the limitations of input views by integrating an aerial perspective to provide a comprehensive global view of road scenes, in contrast to the localized perspective obtained from ground-level vehicle cameras. The aerial perspective can be captured using devices such as drones.

\begin{figure*}
\begin{center}
\includegraphics[width=0.95\linewidth]{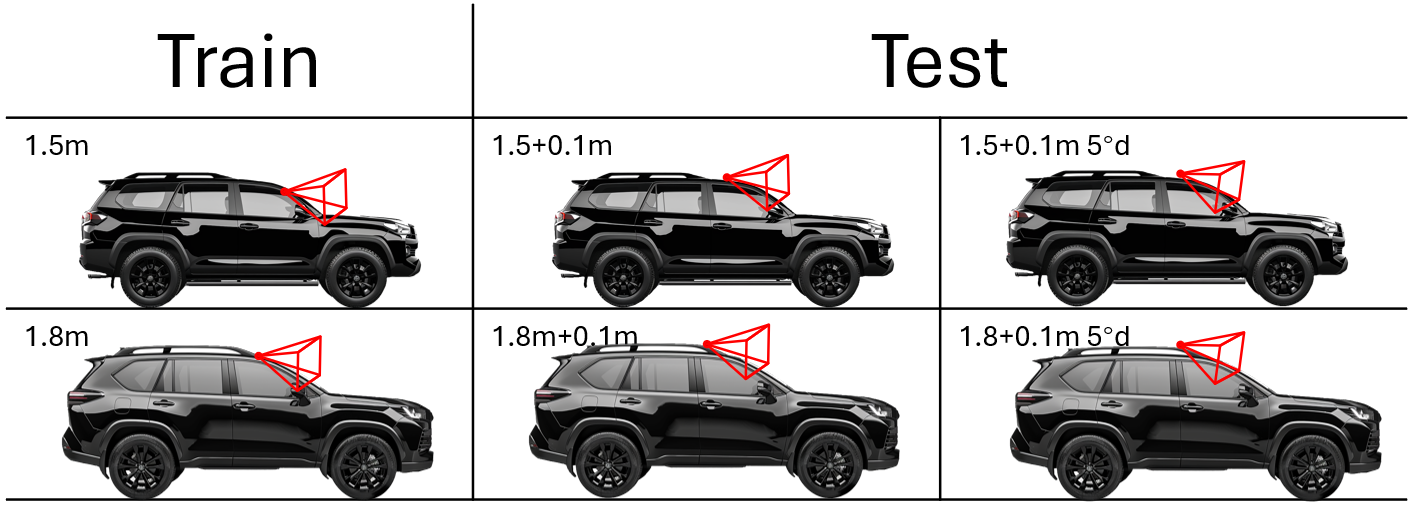}
\end{center}
\vspace{-5mm}
   \caption{General view of the synthesized dataset. 5°d means 5 degrees downward.}
\endcenter
\label{fig:dataset}
\vspace{-5mm}
\end{figure*}

Since it is difficult to capture well-aligned view synthesis ground truth dataset for evaluation in real world, we first create a synthesized dataset (\cref{fig:dataset}) comprising aerial-ground imagery with viewpoints that have similar levels of information granularity across large-scale road scenes. For the aerial perspective, we simulate drone flight trajectories and behavior patterns using AirSim \cite{shah2018airsim}. For the ground-view images, we sample them to simulate the perspective and field of view from onboard cameras on vehicles.

On the other hand, the training process of 3D-GS leads it to an imperfect modeling when it comes to significant different viewpoints, such as aerial and ground views. This limited generalization capability is exacerbated in our specific setting, making simultaneous fitting of aerial and ground data challenging. For instance, the inherent ambiguity in aerial-ground data and the information outside co-view regions dilute useful information. Naively incorporating aerial data into the joint training with ground data adversely affects the convergence of the Gaussian field and compromises rendering quality for NVS when shifting and rotating views in autonomous driving scenes.

To overcome this problem, we introduce a novel uncertainty-aware training paradigm that enables more effective use of aerial imagery and guides 3D-GS to focus on challenging areas where ground data alone may struggle. By excluding irrelevant portions in aerial perspectives, such as the upper floors of buildings, which are less related to street scenes, we successfully mitigate the ambiguity and improve the fidelity of NVS such as view shifting and rotation on street.


The uncertainty is first computed within the ground-view image space through an ensemble-based method, and then projected to the aerial space to assist the training of 3D-GS, which is named as cross-view uncertainty as a new concept. Extensive experiments demonstrate that our uncertainty-aware training paradigm for drone-assisted road scene synthesis out-performs the naive joint training with aerial and ground images, as well as training with only ground images, both quantitatively and qualitatively. Our method offers significant benefits for applications like autonomous driving simulation. To summarize, the contributions of our work include:
\begin{itemize}
    \item We formalize the problem of drone-assisted road scene synthesis and craft a high-quality and appropriate dataset for this new and important problem; 
    \item We propose an uncertainty-aware training strategy and are the first to demonstrate that cross-view uncertainty can facilitate a pixel-weighted training paradigm while prior works use all pixels of images equally for 3D Gaussians' training;
    \item Through extensive experiments and evaluations, we demonstrate notably improved performance on both held-out tests and scenarios involving view shifting and rotation on road scene synthesis. 
\end{itemize}

\section{Related Works}
\label{sec:related work}

\subsection{3D Scene Representation}

As the foundation for 3D computer vision, various 3D representations have been proposed to depict real-world scenes such as point cloud-based representation \cite{fastlio2, murORB2}, voxel-based representation \cite{mao2021voxel,voxelmap}, or implicit representation \cite{park2019deepsdf, mildenhall2021nerf}. Among the implicit representation, NeRF \cite{mildenhall2021nerf} stands out as a groundbreaking neural rendering method that represents 3D scenes as continuous radiance fields parameterized by neural networks, taking coordinates and viewing directions as inputs. With the rise of NeRF, many efforts have been made to enhance its quality and efficiency \cite{yu2021pixelnerf,xu2022sinnerf,deng2022depth, barron2021mip,barron2022mip,barron2023zip,liu2024rip,yuan2024slimmerf,xu2024camera,chen2023nerrf}. Recently, 3D Gaussian splatting (3D-GS) \cite{kerbl20233d} has been proposed as a novel 3D scene representation, utilizing a set of 3D positions, opacity, anisotropic covariance, and spherical harmonic (SH) coefficients to represent a 3D scene. Compared with NeRF, 3D-GS based methods \cite{yu2023mip, lu2023scaffold, cheng2024gaussianpro,song2024sa,yang2024spectrally}, shows superior performance in rendering speed, fidelity, and training time. In this work, we also leverage 3D-GS as the scene representation to resolve the problem of drone-assisted road scene synthesis.

\subsection{Uncertainty Modeling}

Modeling uncertainty has been a long-standing problem in deep learning. Early works usually resolve uncertainty estimation through Bayesian Neural Network (BNN) \cite{neal2012bayesian,neapolitan2004learning}.  However, these methods can be computationally expensive and challenging to implement. Later, dropout-based methods \cite{gal2016dropout,milanes2021monte,tian2022vibus} have emerged as a computationally efficient alternative that adds dropout during inference to estimate uncertainty. Besides, ensemble-based methods \cite{lakshminarayanan2017simple,liu2019accurate} have been proposed to model uncertainty by merging the prediction from multiple independently trained neural networks. As for the field of 3D scene representation, a series of works \cite{shen2021stochastic, shen2022conditional, goli2023bayes,shen2023estimating} have focused on quantifying uncertainty in the prediction of NeRF. For example, S-NeRF \cite{shen2021stochastic}  employ
a probabilistic model to learn a simple distribution over radiance fields, while CF-NeRF \cite{shen2022conditional} learns a distribution over possible radiance fields with latent variable modeling and conditional normalizing flows. With the emergence of 3D-GS,  SGS \cite{savant2024modeling} first addresses uncertainty modeling in 3D-GS, and integrates a variational inference-based method with the rendering pipeline of 3D-GS. CG-SLAM \cite{hu2024cg} also introduce the uncertainty-aware 3D-GS to SLAM. In this work, we introduce a novel cross-view uncertainty training paradigm to facilitate the training of 3D Gaussians on road scenes.

\section{Method}
\label{sec:method}
\begin{figure*}
\begin{center}
\includegraphics[width=0.95\linewidth]{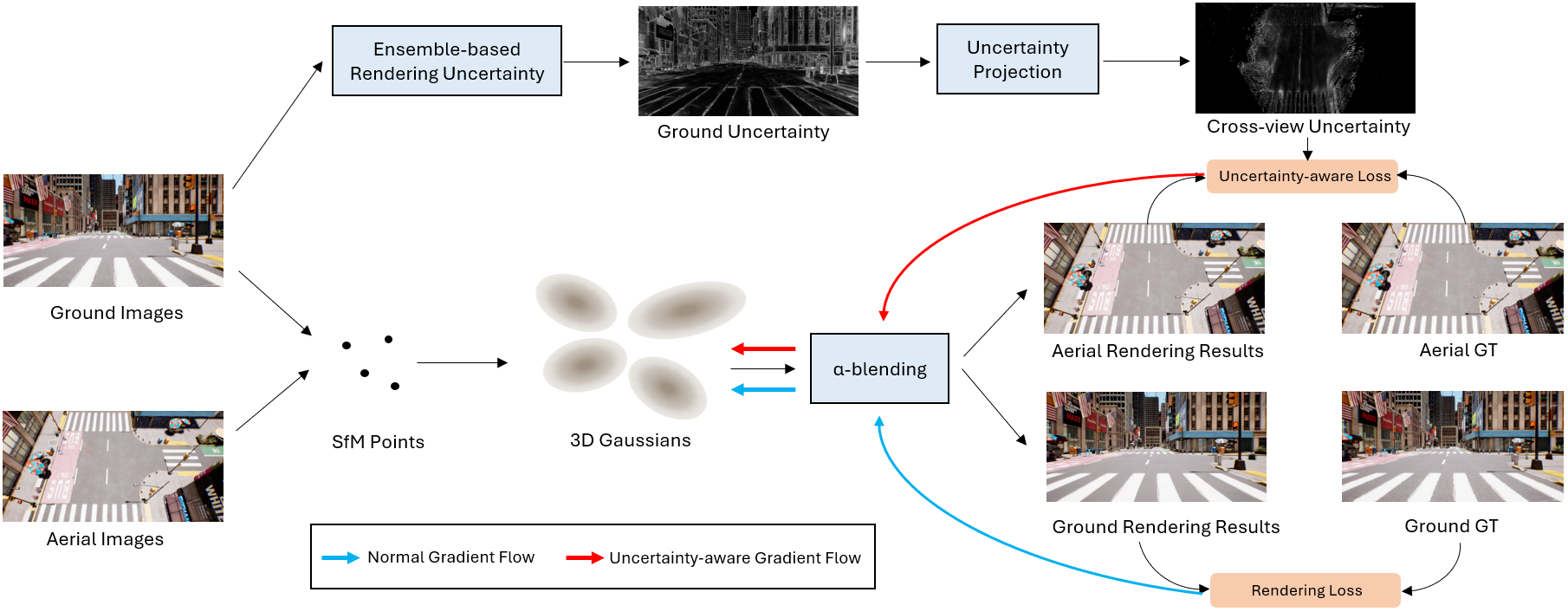}
\end{center}
\vspace{-5mm}
   \caption{Overview of Drone-assisted Road Gaussian Splatting with Cross-view Uncertainty. We first adopt an ensemble-based rendering uncertainty to quantify the learning outcomes of 3D Gaussians on ground images. Next, the ground uncertainty is projected to the air to build the cross-view uncertainty. Subsequently, we introduce the cross-view uncertainty to the training of 3D Gaussians as weight for each pixel of aerial images in the loss function, together with the original rendering loss of 3D-GS for ground images.}
\label{fig:short}
\vspace{-5mm}
\end{figure*}

\cref{fig:short} depicts the overview of our method. In \cref{3-1}, we briefly introduce the basic principles of original 3D-GS. Next, we construct the first drone-assisted road scene dataset in \cref{road scene synthesis}. Then, \cref{3-3} illustrates how we model cross-view uncertainty through an ensemble-based rendering paradigm and an uncertainty projection module. Finally, by incorporating the uncertainty map into the loss function, we can build an uncertainty-aware training module, which facilitates the training of 3D-GS (\cref{3-4}). 


\subsection{Preliminaries}
\label{3-1}

3D-GS \cite{kerbl20233d} represents a 3D scene by a set of differentiable 3D Gaussians, which could be efficiently rendered to images through tile-based rasterization.

Specifically, initialized by a bunch of Structure-from-Motion (SfM) points, each 3D Gaussian is defined as:
\begin{equation}
G(x) = e^{-\frac{1}{2} (x-\mu)^T \Sigma^{-1} (x-\mu)},
\label{gs}
\end{equation}
where $x \in \mathbb{R}^{3 \times 1} $ is a random 3D position in the scene, $\mu \in \mathbb{R}^{3 \times 1}$ stands for the mean vector of the 3D Gaussian, and $\Sigma \in \mathbb{R}^{3 \times 3}$ refers to its covariance matrix. In order to maintain its positive semi-definite, $\Sigma$ is further formulated as $\Sigma=RSS^TR^T$, where $R \in \mathbb{R}^{3 \times 3}$ is the rotation matrix and $S \in \mathbb{R}^{3 \times 3}$  is the scaling matrix. 

To render the Gaussians into the image space, each pixel $p$ is colored by $\alpha$-blending $N$ sorted Gaussians overlapping $p$ as:
\begin{equation}
\label{eq:colorblending}
c(p)=\sum_{i=1}^N c_i \alpha_i \prod_{j=1}^{i-1}\left(1-\alpha_j\right),
\end{equation}
where $\alpha_j$ is calculated by multiplying the 2D Guassian projected from 3D Gaussian $G$ in $p$ with the opacity of $G$, and $c_i$ is the color of $G$. Through the differentiable tile-based rasterizer technique, all attributes of the Gaussians could be learnable and optimized end-to-end via training view reconstruction.

In this work, we utilize Scaffold-GS \cite{lu2023scaffold} as our baseline, as it represents the SOTA among 3D-GS based methods in road scene synthesis tasks. However, we posit that our proposed strategy holds promise for application across other 3D-GS based methods as well.

\subsection{Drone-assisted Road Scene Synthesized Dataset} \label{road scene synthesis} 

Crucial for autonomous driving simulation, high-fidelity view synthesis for large road scenes is often hindered by poor road rendering due to reliance on limited car-view imagery. To address this, we introduce drone-assisted road scene synthesis, leveraging aerial images as an additional input for a better scene reconstruction.

We present a new benchmark for assessing aerial images in large-scale road scene synthesis, featuring both aerial and ground posed images. Using Unreal Engine, we create two high-fidelity scenes to simulate real-world road imagery. AirSim \cite{shah2018airsim} controls drones and vehicles for precise trajectory generation, simulating real scenarios. With the trajectories of drones, we employ AirSim to simulate the camera perspectives and render corresponding image data through the Unreal Engine. For ground-view imagery, we utilize vehicle trajectories to generate forward-facing images. As shown in \cref{fig:dataset}, to replicate real-world driving conditions, we capture front-view ground images at heights of 1.5m and 1.8m, while aerial-view images are collected at the height 20m with the angle of 60° downward from the front view (based on some tests). Additional test data at 1.6m and 1.9m heights evaluate perspective impact. Each scene includes a training set of 315 ground and 351 aerial images, and a test set of 36 ground images, aiming to simulate diverse driving scenarios for a more representative benchmark dataset.

\begin{figure}
\begin{center}
\begin{tabular}{{@{}c@{\hspace{2pt}}c@{\hspace{2pt}}c@{}}}
\bmvaHangBox{\includegraphics[width=4.2cm]{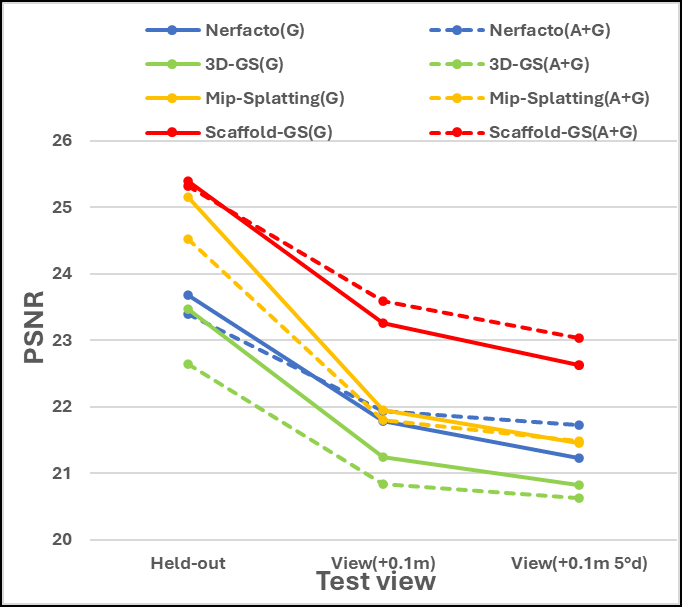}}&
\bmvaHangBox{\includegraphics[width=4.2cm]{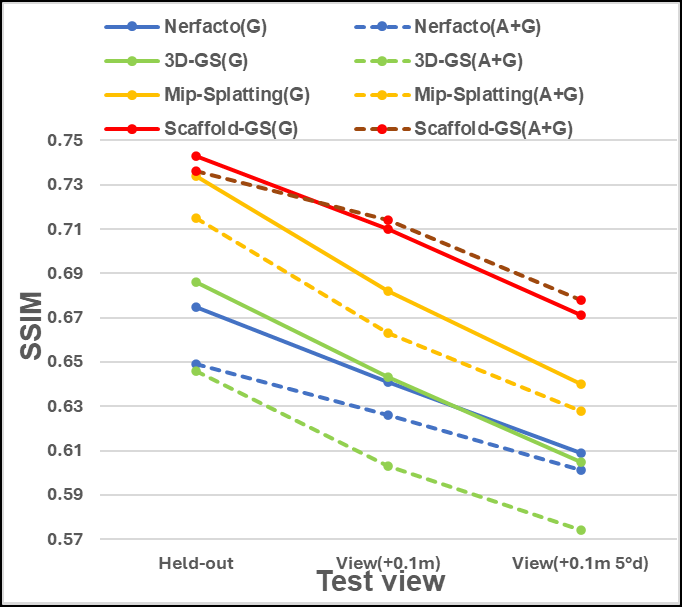}}&
\bmvaHangBox{\includegraphics[width=4.2cm]{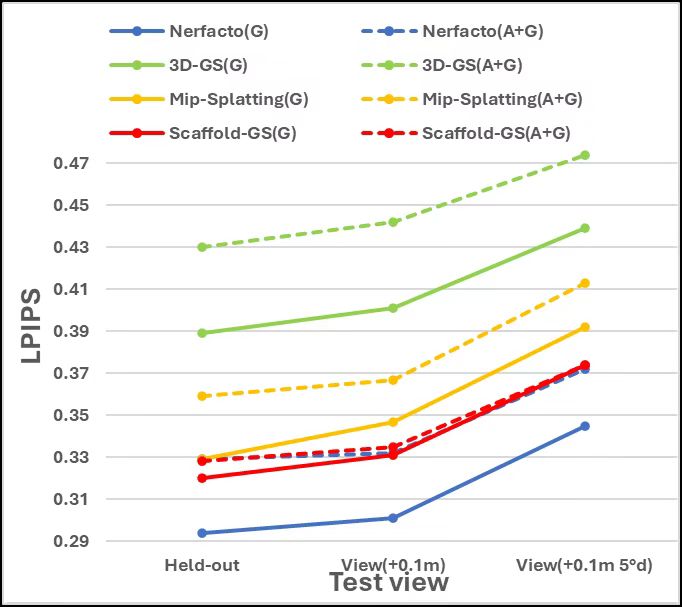}}\\
(a)&(b)&(c)
\end{tabular}
\end{center}
\caption{Results for training with ground or aerial and ground images on various models. (G), (A+G) are training with ground or aerial and ground images.}
\label{fig:Pre-exp}
\vspace{-5mm}
\end{figure}

\subsection{Cross-view Uncertainty Modeling}
\label{3-3}

\textbf{Preliminary Experiments and Motivation.} From the comparison between the dotted and solid lines of different colors in \cref{fig:Pre-exp} (a/b/c), it is clearly that jointly training aerial and ground images mitigates the decline in metrics during road view shifting and rotation compared with merely training with ground images. However, aerial images do not enhance the result on the held-out test of road scene synthesis, as shown by the point in the held-out column of \cref{fig:Pre-exp} (a/b/c). Weighting all pixels from aerial and ground images equally while training will let aerial images have same synthesis priority as road views do for 3D Gaussians. The areas that are non-overlapped with road scene and the areas that have little contribution to the road scene synthesis in the aerial images not only fail to enhance the effectiveness of road reconstruction but also pose more challenges to 3D Gaussians' convergence. This leads to poor rendering quality in the ground perspective when jointly training aerial and ground images. 

\noindent\textbf{Baseline and Implementation.} In order to enhance the rendering result of road views, we attempt to quantify the contribution of each pixel in the aerial image to the road scene synthesis. However, undertaking such a task is very challenging. We decide to approach from a different angle by quantifying the quality of the learning outcomes of ground images to infer the weight to each pixel in the aerial image during Gaussians' training. To conveniently and plausibly compute the learning outcomes, we adopt an ensemble-based rendering uncertainty \cite{lakshminarayanan2017simple} paradigm to quantify the uncertainty of each pixel in the ground imagery. To be more specific, we train an \emph{ensemble} of $M$ gaussian splatting(GS)s initialised from the structure from motion (SfM) generated by ground imagery. By interpreting the ensemble as a uniformly-weighted mixture model, the members' predictions are combined through averaging, and the predictive uncertainty is expressed as the variance over the individual member predictions. With an ensemble of GSs, the expected color of pixel $p$ in a scene is
\begin{equation}
\mu_\text{RGB}(p) = \frac{1}{M}\sum_{k=1}^M c_k(p).
\end{equation}
The predictive uncertainty can be expressed as the variance over the individual member predictions: 
\begin{equation}
    \sigma_\text{RGB}^2(p) = \frac{1}{M}\sum_{k=1}^M \left(\mu(p) - c_k(p)\right)^2.
\end{equation}
$\mu_\text{RGB}$ and $\sigma_\text{RGB}^2$ can be calculated very easily by rendering the $M$ individual RGB images and calculating the mean and variance directly in pixel space. Both will be 3-vectors over the RGB colour channels.

We combine the variances from the colour channels into a single uncertainty value by:
\begin{equation}
    u(p) = \frac{1}{3}\sum_{c \in \{RGB\}} \log(\sigma^2_{\text{RGB}, (c)}(p)+1), 
    \label{eq:sigma_rgb}
\end{equation}
where $\sigma^2_{\text{RGB}, (c)}(p)$ indicates the variance associated with colour channel $c$, $\log$ is the logarithmic transformation to smooth and tighten the values for further normalization process.

\begin{figure*}
\begin{center}
\includegraphics[width=1.0\linewidth]{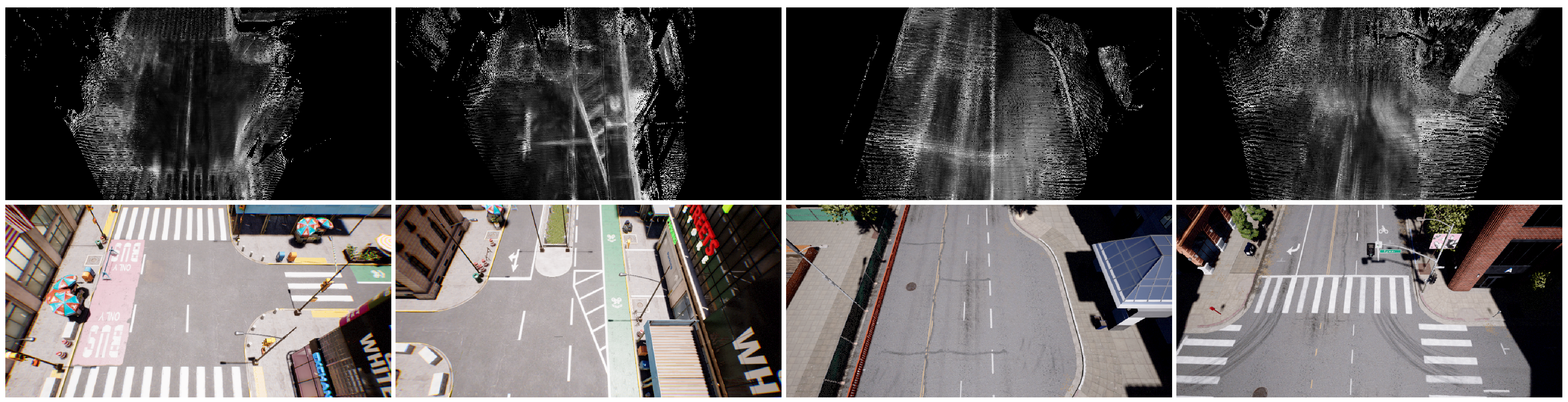}
\end{center}
\vspace{-5mm}
   \caption{The first row shows the visualization of the cross-view uncertainty, and the second row shows the corresponding aerial data.}
\endcenter
\label{fig:uncertainty}
\vspace{-5mm}
\end{figure*}

\noindent\textbf{Cross-view Uncertainty Projection.} To project the uncertainty map from ground-view to aerial-view, we test several methods. Neural field-based methods like NeRF and 3D-GS are prone to overfitting, so neural fields trained with ground uncertainty maps are unable to render high-quality uncertainty in the air. Besides, recently appeared end-to-end dense stereo model—DUSt3R \cite{wang2023dust3r} has set SoTAs on many 3D tasks, which could be used as a 2D-2D pixel matcher between aerial and ground images. In this way, uncertainty maps from ground are projected to air through matches between ground and aerial images, and by averaging the uncertainties at pixels with multiple matches, we build reasonable cross-view uncertainty maps for training. The visualization of the cross-view uncertainty map is shown in \cref{fig:uncertainty}.

\subsection{Uncertainty-aware Training}
\label{3-4}

In this section, we elaborate on how we introduce the cross-view uncertainty map to the training process. $U_k(x)$ presents the uncertainty value on the pixel position $x$ of the $k$-th aerial image. First of all, we normalize all the uncertainty to the range (0, 1) like this:
\begin{equation}
\label{eq:uc}
    U^{\prime}_k = {(\frac{U_k}{max(U_1...U_M)-min(U_1...U_M)})}^{\frac{1}{n}},
\end{equation}
where $n$ refers to a hyperparameter for taking the $n$-th root, with the purpose of enhancing the impact of non-zero values.

Then, we introduce the uncertainty map to the color and SSIM loss as a weight map:
\begin{equation}
    \mathcal{L}_\text{color} = \frac{1}{H  W} \sum_{x=1}^{HW} U^{\prime}(x)|\hat{C}(x) - C(x)|,
\end{equation}
\begin{equation}
    \mathcal{L}_\text{SSIM} = \text{mean}(U^{\prime}(1.0 - \text{SSIM\_MAP}(\hat{C}, C))),
\end{equation}
where $\hat{C}(x)$ and $\hat{C}$ represent ground-truth color, $H$ and $W$ stand for height and width of the image, and $\text{SSIM\_MAP}$ is the structural similarity of the inference and the ground-truth. The final loss function is given by:
\begin{equation}
\label{eq:loss}
    \mathcal{L} = (1.0-\lambda_{\text{SSIM}}) \mathcal{L}_\text{color} + \lambda_{\text{SSIM}}\mathcal{L}_{\text{SSIM}} + \lambda_{\text{vol}}\mathcal{L}_{\text{vol}},
\end{equation}
where $\mathcal{L}_{\text{vol}}$ is the volume regularization used in \cite{lu2023scaffold} to encourages the neural Gaussians to be small with minimal overlapping.

The loss function achieves the cross-view uncertainty-aware training which weights the effects of each pixel of aerial images to better assist in the road scene synthesis.

\section{Experiments}
\subsection{Experimental Setup}
\label{4-1}
\textbf{Dataset and Metrics.} In order to ensure the authenticity of the simulation data,  we use two realistic city scene model, hereafter as New York City (NYC) and San Francisco (SF), from  Kyrylo Sibiriakov \cite{ArtStation} and Tav Shande \cite{ArtStation1} to collect the data as mentioned in \cref{road scene synthesis}. In addition to the $960 \times 480$ aerial and ground images, we also collected $1280 \times 720$ (HD) aerial images for experiments. All models are trained on 1.5m and 1.8m ground images, respectively, and tested at the viewpoint of the front and 5° downward at 0.1 meter above the ground level. All results are measured by three metrics: PSNR, SSIM and LPIPS.

\noindent\textbf{Baseline and Implementation.} Through preliminary experiments among several methods, Scaffold-GS \cite{lu2023scaffold} is selected as the baseline since its outstanding performance. All methods are trained for 900k iterations. Furthermore, we record the results of other SoTA methods in NVS like Nerfacto \cite{tancik2023nerfstudio}, 3D-GS \cite{kerbl20233d} and Mip-Splatting \cite{yu2023mip}. 

For hyperparameters in \cref{eq:loss}, $\lambda_{\text{SSIM}} = 0.2$ and $\lambda_{\text{vol}} = 0.001$ as in \cite{lu2023scaffold}. For $n$ in \cref{eq:uc}, the default value is $6$ and we will discuss it in \cref{4-3}.

\subsection{Main Results}
\label{4-2}
From preliminary experiments (\cref{fig:Pre-exp}), it could be easily concluded that when testing the view shifting and rotation, the metrics of all methods decline. The inclusion of aerial images helps to slow down this trend compared to training the ground data along, indicating that aerial images can provide more perspective-rich information to maintain the rendering ability of the neural field during the view shifting and rotation. However, in the held-out viewpoints testing, training with aerial images performs no better than merely training on ground images, thus failing to demonstrate the superiority of aerial images in terms of view shifting and rotation.

\cref{tab:1} reports comprehensive results of road view synthesis on our datasets. After the implementation of the cross-view uncertainty, the paradigm makes a great progress. The average growth of PSNR on the held-out test set is 0.68 (NYC), and 0.41 (SF) compared with SoTA methods training with ground images. The SSIM and LPIPS also make a significant improvement. When shifting views, the PSNR is about 0.90 (NYC) and 0.80 (SF) more than training with ground images. The SSIM and LPIPS exhibit similar advancement trends. All our results out-performs previous SOTA solutions and \cref{fig:teaser} shows the improvement of our methods on certain details of road scene. In a word, our method not only enhances the representation of GS from ground perspectives but also improves the quality of road scenes synthesis during the view shifting and rotation.

\begin{table}[htbp]
  \centering
   \fontsize{6.5pt}{7.5pt}\selectfont
  \begin{subfigure}
    \centering
        \begin{tabular}{c|ccc|ccc|ccc}
        \toprule
        \multicolumn{1}{c|}{Test set} & \multicolumn{3}{c|}{Held-out} & \multicolumn{3}{c|}{View(+0.1m)} & \multicolumn{3}{c}{View(+0.1m 5°down)} \\
        \multicolumn{1}{c|}{Method/Metrics} & \multicolumn{1}{c}{PSNR $\uparrow$} & \multicolumn{1}{c}{SSIM $\uparrow$} & \multicolumn{1}{c|}{LPIPS $\downarrow$} & \multicolumn{1}{c}{PSNR $\uparrow$} & \multicolumn{1}{c}{SSIM $\uparrow$} & \multicolumn{1}{c|}{LPIPS $\downarrow$} & \multicolumn{1}{c}{PSNR $\uparrow$} & \multicolumn{1}{c}{SSIM $\uparrow$} & \multicolumn{1}{c}{LPIPS $\downarrow$} \\
        \midrule
        \multicolumn{1}{c|}{Nerfacto(G) \cite{tancik2023nerfstudio}} & 23.54 & 0.719 & 0.245 & 20.19 & 0.663 & \textbf{0.258} & 19.69 & 0.632 & 0.300 \\
        \multicolumn{1}{c|}{3D-GS(G) \cite{kerbl20233d}} & 23.71 & 0.706 & 0.363 & 20.54 & 0.688 & 0.346 & 20.01 & 0.646 & 0.387 \\
        \multicolumn{1}{c|}{Mip-Splatting(G) \cite{yu2023mip}} & 25.35 & 0.779 & 0.302 & 20.51 & 0.710  & 0.302 & 20.03 & 0.668 & 0.350 \\
        \midrule
        \multicolumn{1}{c|}{Scaffold-GS(G) \cite{lu2023scaffold}} & 25.64 & 0.790  & 0.265 & 22.19 & 0.746 & 0.281 & 21.55 & 0.705 & 0.326 \\
        \multicolumn{1}{c|}{Scaffold-GS(A+G)} & 25.66 & 0.782 & 0.273 & 22.56 & 0.744 & 0.286 & 22.10  & 0.709 & 0.328 \\
        \multicolumn{1}{c|}{Scaffold-GS(A*+G)} & 25.68 & 0.784 & 0.274 & 22.91 & 0.751 & 0.284 & 22.38 & 0.715 & 0.326 \\
        \multicolumn{1}{c|}{Ours} & \textbf{26.32} & \textbf{0.802} & \textbf{0.244} & \textbf{23.11} & \textbf{0.766} & \textbf{0.258} & \textbf{22.49} & \textbf{0.725} & \textbf{0.303} \\
        \bottomrule
        \end{tabular}%
\end{subfigure}
\hspace{10pt} (a)

\fontsize{6.5pt}{7.5pt}\selectfont
\begin{subfigure}
\centering
    \begin{tabular}{c|ccc|ccc|ccc}
    \toprule
    \multicolumn{1}{c|}{Test set} & \multicolumn{3}{c|}{Held-out} & \multicolumn{3}{c|}{View(+0.1m)} & \multicolumn{3}{c}{View(+0.1m 5°down)} \\
    \multicolumn{1}{c|}{Method/Metrics} & \multicolumn{1}{c}{PSNR $\uparrow$} & \multicolumn{1}{c}{SSIM $\uparrow$} & \multicolumn{1}{c|}{LPIPS $\downarrow$} & \multicolumn{1}{c}{PSNR $\uparrow$} & \multicolumn{1}{c}{SSIM $\uparrow$} & \multicolumn{1}{c|}{LPIPS $\downarrow$} & \multicolumn{1}{c}{PSNR $\uparrow$} & \multicolumn{1}{c}{SSIM $\uparrow$} & \multicolumn{1}{c}{LPIPS $\downarrow$} \\
    \midrule
    \multicolumn{1}{c|}{Nerfacto(G) \cite{tancik2023nerfstudio}} & 23.82 & 0.631 & 0.344 & 23.38 & 0.618 & 0.344 & 22.77 & 0.587 & 0.421 \\
    \multicolumn{1}{c|}{3D-GS(G) \cite{kerbl20233d}} & 23.22 & 0.630  & 0.449 & 21.96 & 0.597 & 0.457 & 21.64 & 0.564 & 0.492 \\
    \multicolumn{1}{c|}{Mip-Splatting(G) \cite{yu2023mip}} & 24.95 & 0.690  & 0.381 & 23.38 & 0.654 & 0.391 & 22.90 & 0.613 & 0.434 \\
    \midrule
    \multicolumn{1}{c|}{Scaffold-GS(G) \cite{lu2023scaffold}} & 25.16 & 0.697 & 0.375 & 24.37 & 0.675 & 0.386 & 23.69 & 0.637 & 0.423 \\
    \multicolumn{1}{c|}{Scaffold-GS(A+G)} & 24.98 & 0.691 & 0.383 & 24.63 & 0.683 & 0.384 & 23.97 & 0.647 & 0.421 \\
    \multicolumn{1}{c|}{Scaffold-GS(A*+G)} & 24.99 & 0.689 & 0.386 & 24.70 & 0.684 & 0.385 & 23.88 & 0.649 & 0.422 \\
    \multicolumn{1}{c|}{Ours} & \textbf{25.57} & \textbf{0.723} & \textbf{0.337} & \textbf{25.18} & \textbf{0.715} & \textbf{0.338} & \textbf{24.55} & \textbf{0.678} & \textbf{0.376} \\
    \bottomrule
    \end{tabular}
\end{subfigure}
\hspace{10pt} (b)
\vspace{-3mm}
\hfill
\caption{Results on NYC (a) and SF (b). A* is HD aerial images. (G), (A+G) are training with ground or aerial and ground images.} 
\label{tab:1}
\vspace{-5mm}
\end{table}

\subsection{Ablation studies}
\label{4-3}
\begin{figure*}
\begin{center}
\includegraphics[width=1.0\linewidth]{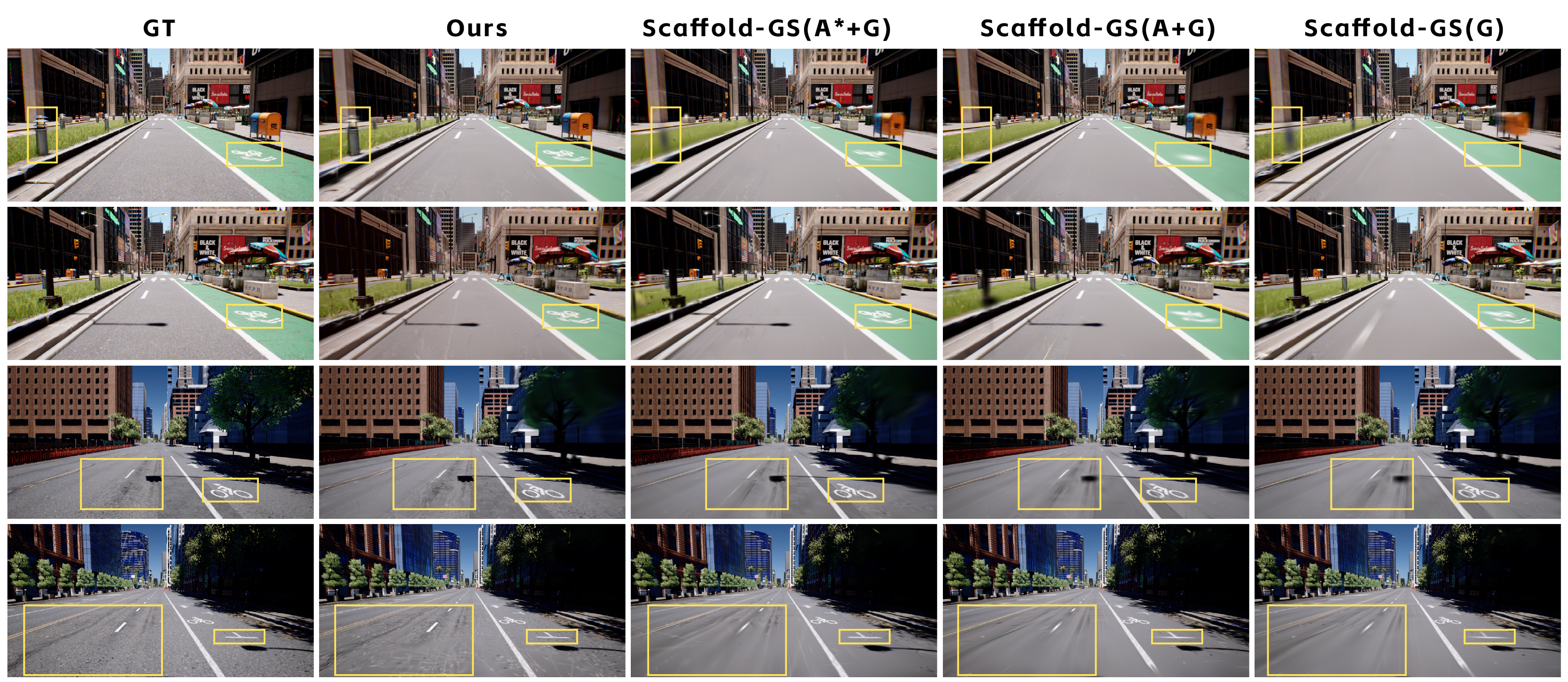}
\end{center}
\vspace{-5mm}
   \caption{Rendering results for the ablation study of cross-view uncertainty. NYC: 1.6m (row 1), 1.6m 5° down (row2); SF: 1.9m (row 3), 1.9m 5° down (row4). A* is HD aerial images. (G), (A+G) are training with ground or aerial and ground images.}
\label{fig:result}
\vspace{-5mm}
\end{figure*}

\textbf{Efficacy of Cross-view Uncertainty.}
Compared with equally training all aerial and ground images \cref{tab:1}, the cross-view uncertainty-aware training achieves a 0.66 (NYC) and 0.59 (SF) increase in PSNR on held-out test set, and about 0.47 (NYC) and 0.57 (SF) when the view shifting and rotation. Moreover, our method also reverses the adverse effects of the joint training on SSIM and LPIPS, resulting in improvements in both metrics. It is also very impressive that our method performs even better than using HD aerial data when the view shifting and rotation. This reflects the effective utilization of aerial data in road scene synthesis. From \cref{fig:result}, it is clear that our method not only contributes to the rendering effect of road textures but also enhances the clarity of roadside obstacles, lane markings and ground signs, which will greatly aid in autonomous driving simulation.

\noindent\textbf{Efficacy of Hyperparameter $n$.}
\cref{tab:2} presents the experimental results for different values of $n$ in \cref{eq:uc}. When $n$ is set to 1 (i.e., no $n$), there is no significant improvement in the metrics compared to equally training on all aerial and ground images. However, as $n$ increases to 2 or greater, the metrics improve with the increment of $n$, and the results become stable when $n\geq6$. This indicates that when $n\geq6$, the potential of aerial imagery is fully realized.

\begin{table}[htbp]
  \centering
  \fontsize{6.5pt}{7.5pt}\selectfont
    \begin{tabular}{c|ccc|ccc|ccc}
    \toprule
    \multicolumn{1}{c|}{Test set} & \multicolumn{3}{c|}{Held-out} & \multicolumn{3}{c|}{View(+0.1m)} & \multicolumn{3}{c}{View(+0.1m 5°down)} \\
    \multicolumn{1}{c|}{{$n$}/Metrics} & \multicolumn{1}{c}{PSNR $\uparrow$} & \multicolumn{1}{c}{SSIM $\uparrow$} & \multicolumn{1}{c|}{LPIPS $\downarrow$} & \multicolumn{1}{c}{PSNR $\uparrow$} & \multicolumn{1}{c}{SSIM $\uparrow$} & \multicolumn{1}{c|}{LPIPS $\downarrow$} & \multicolumn{1}{c}{PSNR $\uparrow$} & \multicolumn{1}{c}{SSIM $\uparrow$} & \multicolumn{1}{c}{LPIPS $\downarrow$} \\
    \midrule
    \multicolumn{1}{c|}{1} & 25.64 & 0.748 & 0.310  & 23.71 & 0.723 & 0.320  & 23.11 & 0.685 & 0.360 \\
    \multicolumn{1}{c|}{2} & 25.72 & 0.754 & 0.301 & 23.83 & 0.729 & 0.309 & 23.24 & 0.690  & 0.353 \\
    \multicolumn{1}{c|}{3} & 25.80 & 0.758 & 0.298 & 23.93 & 0.733 & 0.305 & 23.39 & 0.694 & 0.347 \\
    \multicolumn{1}{c|}{4} & 25.85 & 0.759 & 0.296 & 24.04 & 0.735 & 0.303 & \cellcolor{tabthird}23.47 & 0.698 & 0.344 \\
    \multicolumn{1}{c|}{6} & \cellcolor{tabfirst}25.94 & \cellcolor{tabsecond}0.762 & \cellcolor{tabsecond}0.291 & \cellcolor{tabsecond}24.14 & \cellcolor{tabsecond}0.741 & \cellcolor{tabsecond}0.298 & \cellcolor{tabsecond}23.52 & \cellcolor{tabsecond}0.701 & \cellcolor{tabsecond}0.339 \\
    \multicolumn{1}{c|}{8} & \cellcolor{tabthird}25.89 & \cellcolor{tabthird}0.761 & \cellcolor{tabsecond}0.291 & \cellcolor{tabthird}24.13 & \cellcolor{tabthird}0.739 & \cellcolor{tabsecond}0.298 & \cellcolor{tabthird}23.47 & \cellcolor{tabthird}0.700   & \cellcolor{tabsecond}0.339 \\
    \multicolumn{1}{c|}{10} & \cellcolor{tabsecond}25.91 & \cellcolor{tabfirst}0.763 & \cellcolor{tabfirst}0.290  & \cellcolor{tabfirst}24.20 & \cellcolor{tabfirst}0.742 & \cellcolor{tabfirst}0.296 & \cellcolor{tabfirst}23.55 & \cellcolor{tabfirst}0.704 & \cellcolor{tabfirst}0.337 \\
    \bottomrule
    \end{tabular}%
    \vspace{4pt} 
  \caption{The results for the ablation study of hyperparameter $n$. $n$ is for taking the $n$-th root to the uncertainty map. Metrics are averaged over testing on two datasets.}
  \label{tab:2}%
\vspace{-6mm}
\end{table}%

\section{Conclusion}
In this work, we propose a novel drone assisted road Gaussian Splatting with cross-view uncertainty. To use the global information from images in drones' view to assist ground-view training, we are the first to introduce the cross-view uncertainty into the 3D-GS based model for weighting pixels in aerial images during training. This method reduces the impact of superfluous aerial information and effectively utilizes aerial images for road scene synthesis. From the experimental results, we achieve SoTA on two high-fidelity synthesized datasets. Our method enhances various metrics for held-out ground view synthesis while maintaining the robustness of aerial-ground training during the view shifting and rotation. The superiority of the method shows a great potential for the improvement of autonomous driving simulation in the near future. 

\section*{Acknowledgement} 
This research is sponsored by Tsinghua-Toyota Joint Research Fund (20223930097).

\bibliography{egbib}
\end{document}